\documentclass[10pt,twocolumn,letterpaper]{article}

\usepackage{cvpr}

\usepackage{customization}

\usepackage{enumitem}
\setlist{nolistsep}
\usepackage{caption} 
\usepackage[dvipsnames]{xcolor}

\usepackage[accsupp]{axessibility}  

\definecolor{cvprblue}{rgb}{0.21,0.49,0.74}
\usepackage[pagebackref,breaklinks,colorlinks,allcolors=cvprblue]{hyperref}

\def\paperID{15} 
\def\confName{CVPRW}
\def\confYear{2025}

\title{Improving Open-World Object Localization by Discovering Background}

\author{Ashish Singh\thanks{Ashish did most of this work while an intern at MERL.}\\
Univ. of Mass.-Amherst\\
{\tt\small ashishsingh@cs.umass.edu}
\and
Michael J. Jones\\
Mitsubishi Electric Research Labs\\
{\tt\small mjones@merl.com}
\and
Kuan-Chuan Peng\\
Mitsubishi Electric Research Labs\\
{\tt\small kpeng@merl.com}
\and
Anoop Cherian\\
Mitsubishi Electric Research Labs\\
{\tt\small cherian@merl.com}
\and
Moitreya Chatterjee\\
Mitsubishi Electric Research Labs\\
{\tt\small chatterjee@merl.com}
\and
Erik Learned-Miller\\
Univ.of Mass.-Amherst\\
{\tt\small elm@cs.umass.edu}
}

\begin{document}
\maketitle

\begin{abstract}
Our work addresses the problem of learning to localize objects in an open-world setting, \ie, given the bounding box information of a limited number of object classes during training, the goal is to localize all objects, belonging to both the training and unseen classes in an image, during inference. Towards this end, recent work in this area has focused on improving the characterization of objects either explicitly by  proposing new objective functions (localization quality) or implicitly using object-centric auxiliary-information, such as depth information, pixel/region affinity map \etc In this work, we address this problem by incorporating background information to guide the learning of the notion of objectness. Specifically, we propose a novel framework to discover background regions in an image and train an object proposal network to not detect any objects in these regions. We formulate the background discovery task as that of identifying image regions that are not discriminative, \ie, those that are redundant and constitute low information content. We conduct experiments on standard benchmarks to showcase the effectiveness of our proposed approach and observe 
significant improvements over the previous state-of-the-art approaches for this task.
\end{abstract}    
\section{Introduction}
\label{sec:intro}

\begin{figure}[t]
\centering
\includegraphics[width=0.85\linewidth]{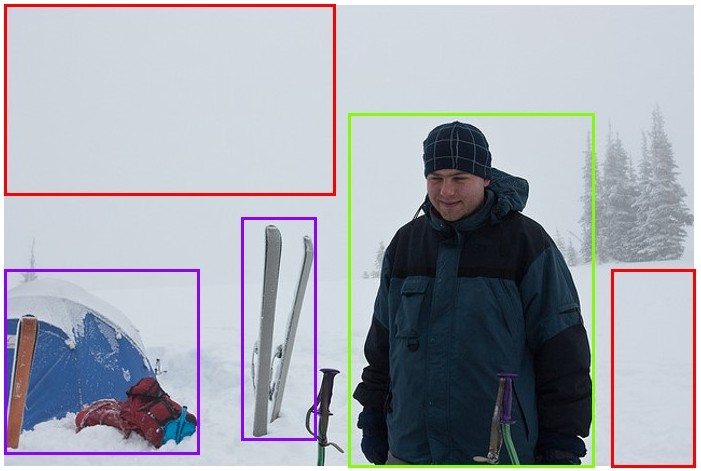}
\caption{An example of a training image showing a ground truth annotated object (green box), unlabeled objects that are not in the known training classes (purple boxes), and clear non-object/background regions (red boxes) that are automatically classified as non-object regions by \ourso and used in training.}
\label{fig:teaser}
\end{figure}

Open-world object localization (OWOL) refers to the task of localizing all objects, including novel object instances in a given image by learning from a limited set of predefined object classes. Unlike the traditional closed-world object localization setting, where the task is to localize object classes only seen during training, OWOL is a more realistic and challenging problem that evaluates generalization performance of the learned model to novel object classes. 

Learning to localize object instances, not seen during training time, is important for many downstream applications like autonomous agents, robotics, and self-driving cars. As a result, there has been a growing interest in developing methods for object detection/localization in an open-world setting over the last few years \cite{kim2021oln, wang2022open}. Previous works in OWOL have primarily focused on improving objectness learning. Specifically, Kim \etal \cite{kim2021oln} proposed incorporating localization-quality as a loss function to learn a more generalizable object representation, whereas, Wang \etal \cite{wang2022open} explicitly enriches the training data by synthesizing pseudo-annotations for unannotated regions. While the above approaches show improvement in localizing unseen object classes, their performance is still limited. Specifically, we observe that during inference, either: \textbf{(1)} novel objects with features like shape, color, textures, \etc not observed in known classes are not detected, or \textbf{(2)} non-object regions are erroneously detected as objects due to noisy pseudo-annotations during model training. These observations allude to the challenging issue of developing and learning a generalizable notion of objectness. Learning the objectness of a region is non-trivial and most assumptions developed to characterize objectness do not generalize to unseen environments. In this work, we tackle this challenge by exploring a complementary source of information, \ie, non-objectness of a region. We show that identifying non-object regions and incorporating them during the learning process significantly improves the localization of unseen object classes and the generalization to novel domains on various challenging benchmarks.

In particular, we propose to improve open-world object localization by incorporating non-objectness information during the learning process as illustrated in Figure \ref{fig:teaser}. We first model the non-objectness information by constructing a codebook comprised of embeddings of exemplar non-object patches, extracted from a pre-trained model. This codebook is then used to identify seed anchors representing non-object regions while training the object localization method. This provides additional self-supervision in training the localization method along with labeled information from known classes. 

Our evaluations show that by incorporating our non-object model into the object localization pipeline, our proposed method, Background-aware Open-World Localizer (\ours), significantly improves over previous methods on multiple benchmarks for OWOL. Specifically, we surpass the previous baselines by 4.0\% on the AR@100 measure for detecting non-VOC classes when the training data consist of 20 PASCAL-VOC \cite{everingham2010pascal} classes. Furthermore, we show a domain generalization improvement of 2.0\% on the AR@100 {measure for} open-world localization in the wild, when evaluating a model trained with the COCO dataset \cite{lin2014microsoft} on the ADE20K dataset \cite{zhou2019semantic}.  
Going forward, we sometimes use the term, "background", in an informal way to mean non-object regions.

Overall, in this work, we make the following contributions:
\begin{enumerate}[label=\arabic*., leftmargin=*, noitemsep,topsep=0pt]
    \setlength\itemsep{-0.0em}
    \item We propose \ours, a novel, open-world object localization method based on background-awareness.
    \item We formalize the notion of non-objectness of image regions and propose a novel, unsupervised method for discovering non-object regions with high precision. 
    \item Our extensive evaluation shows that incorporating non-objectness information in \ours leads to improved recall in detecting unseen objects, outperforming competing methods for this task.
\end{enumerate}

\section{Related Work}
\label{sec:related_work}
In this section, we discuss related prior works and delineate the differences from our proposed approach.

\textbf{Open-World Object Localization} addresses the problem of localizing objects in images under the open-world setting. Early methods in object localization \cite{alexe2012measuring, arbelaez2014multiscale, endres2010category, manen2013prime, zitnick2014edge} adopted a bottom-up approach to creating a relatively small set of candidate bounding boxes that covered the objects in an image. These methods utilized hand-crafted heuristics based on raw image information like color, edge, and pixel-contrast to characterize objectness \cite{uijlings2013selective, zitnick2014edge}. In contrast, learning based methods \cite{ren2015faster, kuo2015deepbox, li2019zoom, o2015learning, wang2019region} adopt a top-down approach, where the model is trained end-to-end with objectness defined by a set of training categories in the dataset. These learning-based methods show considerable improvement over heuristics-based approaches in terms of precision and recall when localizing object classes from the training set. However, they struggle to localize novel object categories. In particular, these methods incorrectly assume any unannotated region of a training image to be a non-object region and thus suppress detection of novel objects at inference time. To resolve this issue, Kim \etal \cite{kim2021oln} proposed to modify Faster-RCNN \cite{ren2015faster} by replacing classification heads with class-agnostic objectness heads which are trained using base-class annotations, thus avoiding any incorrect assumptions about background. Saito \etal \cite{saito2022learning} proposed to instead crop labeled object regions in training images and paste them into images not containing any unlabeled objects to avoid the faulty assumption that unlabeled regions of training images do not contain any objects. While the above methods focus on mitigating the background bias, another family of methods have focused on improving object localization recall by pseudo-annotating novel objects. In particular, Wang \etal~\cite{wang2022open} learns a model to predict pairwise affinity to segment an image. Prediction from this model is then used to generate new box annotations, which is then used to train a detection model along with the original ground truth base class annotations as supervision. 

Our work proposes a complementary approach, where instead of completely suppressing background or pseudo-labeling novel objects, we identify regions in training images which we can categorize as non-objects with high precision. Our experiments show that using non-object supervision for training the detector is better than completely disregarding it and avoids errors introduced by noisy pseudo-annotations.

\paragraph{Unsupervised Background Estimation:} Most works for estimating background in images fall under the paradigm of either: (i) saliency detection \cite{borji2019salient}, or (ii) figure-ground separation \cite{yin2009shape, yu2001hierarchical, yu2003object, ren2006figure}, where the primary objective is to detect the salient foreground regions. Early attempts to solve the foreground/background separation task focused on thresholding or binary clustering on the color, depth, edge or other hand-crafted features \cite{jones2002statistical, wong2000color, yin2009shape, ren2006figure}. 
Recent methods in unsupervised object localization, a new formulation similar to figure-ground segmentation  \cite{simeoni2024unsupervised}, in particular, utilize self-supervised models \cite{oquab2023dinov2, caron2021emerging} to estimate salient objects. Methods like LOST~\cite{simeoni2021localizing} and TokenCut~\cite{wang2023tokencut} used deep features from pre-trained vision transformers (ViTs)~\cite{caron2021emerging}. LOST defined background regions as the patches that were most correlated with the whole image, while TokenCut investigated the effectiveness of applying spectral clustering \cite{von2007tutorial} to self-supervised ViT features. Both of the above method make strong assumptions in defining objectness. As a result, they don't generalize well to images from different domains. In comparison, FOUND~\cite{simeoni2023unsupervised} proposed discovering the background as the primary step, with the objective of discovering foreground objects. Specifically, they first identify seed patch token with least attention value. Then background is defined as all the patches with similar representation as the seed patch. They further improve the initial coarse background mask by refining it using a lightweight segmentation head. 

While our method shares commonality with the above methods with respect to the use of self-supervised models for extraction of patch features, we differ from the above approaches with respect to our primary objective. In particular, in these prior methods, the main goal is to discover objects in the foreground region, and background estimation is a by-product. Thus they focus more on improving the precision of object discovery. Our goal on the other hand, is to improve novel unseen object localization using background as a supervisory signal. Thus for our problem statement, we want to categorize an image region as background/non-object with high confidence. In this regard, the focus of our unsupervised background estimation method is to achieve high precision with moderate recall, which is different from the above approaches.  

\section{Method}
\label{sec:methodology}

Our goal is to improve OWOL by incorporating non-objectness cues. Specifically, our proposed method \ours is comprised of two main steps: (i) \textbf{Modeling non-object information:} where we construct a codebook containing embeddings of exemplar non-object patches, extracted from a pre-trained model, and (ii) \textbf{Background-aware training:} Where we train an open-world localization model with additional self-supervision using non-object seed anchors discovered using our computed non-object codebook. Fig.~\ref{fig:model} shows an overview of our method.

\subsection{Motivation}
\label{sec:mainidea}

For the object localization task, we are provided with a training dataset $D$ consisting of bounding box annotations $\{b_i\}$ from a pre-defined list $S$ of  known base classes. The goal for the object localization task (also commonly known as object proposal generation) is to generate bounding boxes on unseen images, localizing objects from the known base classes. As such, the primary objective adopted to train these models can be divided into two parts: (i) Classification objective $\cL_{cls}$: which trains the model to categorize a region as `object' or `background,' and (ii) Bounding box regression objective $\cL_{reg}$: which helps the model learn to accurately estimate the coordinates of a bounding box. Formally, given a training image $I$, the training loss can be formulated as:
\begin{equation}
\cL(I) = \frac{1}{|A|} \sum_{i \in A} \cL_{cls}(c_i, \ c^*_i) + \frac{1}{|A_K|} \sum_{i \in A_K} \cL_{reg}(b_i, \ b^*_i), \tag{1}
\label{eq:L}
\end{equation}
where $\{c_i, \: b_i\}$ are the predicted label and bounding box coordinates, $\{c^*_i, \: b^*_i\}$ are the corresponding ground truths, $A$ is the set of all candidate anchor boxes, $A_K$ is the set of candidate anchor boxes with ground truth label $c^*_i = 1$. Here, $c^*_i \in C = \{ 0, 1\}$, where $c^*_i = 1$ only when the anchor $i$ can be associated with an annotated object bounding box from the known classes $S$ (based on pre-defined intersection-over-union thresholds) and $c^*_i = 0$ otherwise. 

Thus, every candidate anchor box $A_i$ not associated with an annotated object bounding box is automatically labeled as a background/non-object region. This design choice made sense for traditional methods \cite{ren2015faster}, where the model was trained under the closed-world assumption, \ie, at the testing time, their generalization is evaluated by detecting the objects from the known base classes. However, under the open-world setting, where the task is to localize every object in the image, which can belong to an unknown novel class $u \in U$, the training objective Eq.~\ref{eq:L} can lead to poor test time performance. Specifically, under the training loss in Eq.~\ref{eq:L}, an unannotated object will be classified as `background,' resulting in incorrect categorization of unknown novel object regions as `non-objects.'

To avoid the above issue of false negative detections, Kim \etal~\cite{kim2021oln} proposed to replace $\cL_{cls}$ in Eq.~\ref{eq:L} with the objectness score prediction loss $\cL_{obj}$, yielding:
\fontsize{9pt}{9pt}
\begin{equation}
\cL_{OLN}(I) = \frac{1}{|A_K|} \sum_{i \in A_K} \cL_{reg}(b_i, \: b^*_i) + \frac{1}{|A_K|} \sum_{i \in A_K} \cL_{obj}(o_i, \: o^*_i),
\tag{2}
\label{eq:L_oln}
\end{equation}
\normalsize
where $o_i$ and $o^*_i$ are the predicted and ground truth objectness score with respect to an anchor box $A_i$. Here objectness score can refer to any metric \cite{tian2020fcos, rezatofighi2019generalized} that can measure bounding box localization quality with respect to ground-truth bounding boxes. Thus, in their formulation, Kim \etal~\cite{kim2021oln} removes any dependency of predicting `background' regions. In other words, `background' regions are simply ignored during training.

While the above loss modification does improve performance over traditional methods with respect to localizing unknown novel objects, it underutilizes the supervision from non-object candidate anchor boxes. Specifically, Kim \etal~\cite{kim2021oln} selects the candidate anchor boxes set $A_K$, by randomly sampling $|A_K|$ anchors with  Intersection over Union (IoU) larger than $0.3$ with the matched ground-truth boxes. While this ensures that an anchor box localized at an unknown novel object region is not incorrectly labeled as non-object, it reduces the diversity of anchor boxes in the candidate set, resulting in a reduction of the range of loss values. This issue can be potentially resolved if we knew apriori which regions correspond to non-objects. To this end, we propose the following modification to Eq.~\ref{eq:L_oln}:
\begingroup
\setlength{\abovedisplayskip}{6pt}
\setlength{\belowdisplayskip}{6pt}
\begin{equation}
\begin{split}
\cL_{BOWL}(I) &= \frac{1}{|A_K|} \sum_{i \in A_K} \cL_{reg}(b_i, \  b^*_i) \\
&+ \frac{1}{|A_K|+|A_B|} \sum_{i \in A_K \cup A_B} \cL_{obj}(o_i, \ o^*_i),
\end{split}
\tag{3}
\end{equation}
\endgroup
where $A_B$ refers to the set of non-object anchor boxes. Concretely, we propose to improve the diversity of sampled anchor boxes for objectness score prediction loss by sampling anchor boxes corresponding to non-object regions. Unlike previous methods, we first identify highly likely non-object regions in a given image, to ensure that we do not incorrectly label unknown novel object anchors as non-objects.


\subsection{ Modeling non-object patches in images }
\label{sec:non-objects}

\begin{figure*}[t]
\centering
\includegraphics[width=0.8\linewidth]{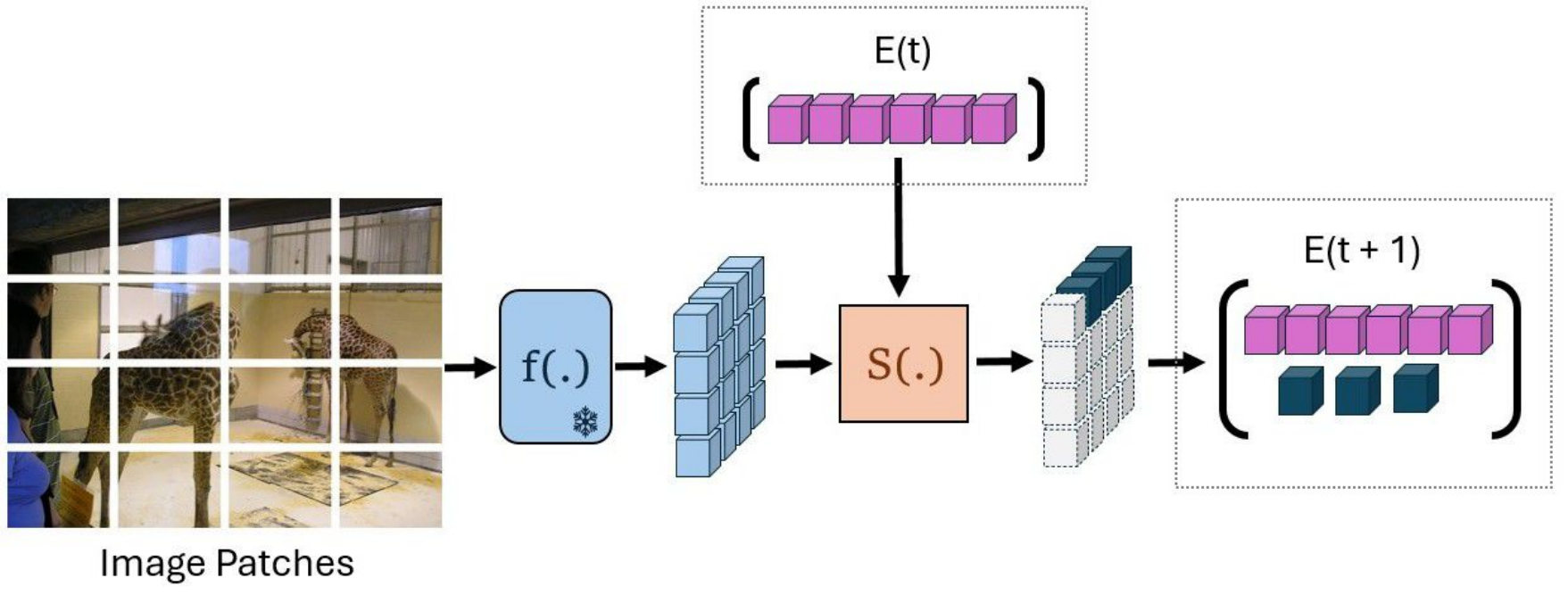}
\caption{Exemplar selection works by first splitting the $t+1^{st}$ image into fixed-size patches and computing a feature embedding (using DINO, represented by the function, $f$) for each patch (represented by the grid of light blue cubes).  Each patch is compared to the existing exemplars from the first $t$ images, $E(t)$, using the similarity function $S$.  $E(0)$ is initialed to the empty set. 
 Patch embeddings that have below threshold similarity to all exemplars in $E(t)$ (represented by dark blue cubes) are then added to the exemplar set to yield $E(t+1)$.  This repeats until all training images are processed.  In addition, each exemplar in $E(t)$ maintains a count of the number of times it was the most similar exemplar to a training patch embedding to keep track of each exemplar's cluster size.}
\label{fig:exemplars}
\end{figure*}

Given an arbitrary training set of unlabeled images, our goal is to identify image regions that are very unlikely to contain objects.  Our model of non-object regions will consist of a representative set (called an exemplar set) of embeddings of non-object patches that cover a large percentage of all non-object patches in the training images.  Note that our non-object model does not need to include every possible non-object region.  Our goal is to learn a representative set of regions that with high probability are all non-object regions.  When we use our non-object model to train the object localization network as described in Section \ref{sec:mainidea} there may still be regions which are ignored because they are neither high-probability non-object regions nor labeled object regions.

To find non-object regions from an unlabeled set of training images, we observe that patches of training images that occur frequently are almost always non-object regions.  For example, patches of sky, grass, concrete, brick, and white walls are all examples of non-object regions that occur frequently in many different images. This idea is consistent with the idea introduced in Sing \etal~\cite{singh2012unsupervised} that object patches are {\em representative} (meaning they occur often enough in the visual world) and {\em discriminative} (meaning they are different enough from the rest of the visual world).  Since we are trying to model non-object patches, we are looking for patches that are representative but not discriminative. By clustering all patches of an unlabeled training set of images and taking a representative patch from the largest clusters, we will have our exemplar set of non-object regions.

Given a set of unlabeled images $\mathcal{I} = \{I_1, \:...\:,\: I_N\}$ where $I_i \in \mathcal{R}^{W \times H \times 3}$, we first compute patches by dividing each image into $M$ square patches of size $S \times S$. Given the set of image patches $ \mathcal{P} = \{P_1, \: ...\: ,\: P_{N \times M}\} $ where $P_i \in \mathcal{R}^{S \times S \times 3}$, we compute features of each patch using a self-supervised pretrained neural network (we use DINO-ViT\citep{caron2021emerging} in our experiments). Each patch $P_i$ is represented by a $d$-dimensional feature embedding $\phi_{P_i}$ computed using the pretrained model $f$, i.e. $f(P_i) = \phi_{P_i}$. The choice of a self-supervised pretrained model ensures that the feature representations are not biased towards any specific object category, allowing for a more general representation of patches. 

Given the set of patch features, we construct an exemplar set $E$ of patch features using a greedy nearest neighbor method.  In particular, for each patch $P_i \in \mathcal{P}$, we compute:
\begin{equation}
    s_{\text{max}}(P_i, \ E) = \max_{e \in E}  \frac{\phi_{P_i} \cdot \phi_{e}}{\left\| \phi_{P_i}\right\| _{2}\left\| \phi_{e}\right\| _{2}}
\end{equation}

A patch $P_i$ is added to $E$ if:
\begin{equation}
    s_{\text{max}}(P_i, \ E) \; \; < \; \; \lambda,
\end{equation}
where $\lambda$ is a distance threshold controlling the diversity of the exemplar set.  Figure \ref{fig:exemplars} illustrates the exemplar selection algorithm.  This approach avoids the computational overhead of traditional clustering methods while still ensuring that the exemplar set captures the diverse range of non-object regions present in the dataset.

As described in Figure \ref{fig:exemplars}, a count of the  number of times each exemplar was the most similar exemplar to a training patch embedding is kept.  These counts give the size of each exemplar cluster.  As explained in Section \ref{sec:non-objects}, the largest clusters correspond to non-object patches with high probability.  Our method simply uses the $N$ clusters with the largest counts as the non-object exemplars in our background model.   

It is important to note that the non-object model is intended to be learned only once and to be general enough to then be used in conjunction with learning an open-world localization network on any object detection dataset.

\subsection{Background-Aware Open-World Localization}

\begin{figure*}[t]
\centering
\includegraphics[width=0.85\linewidth]{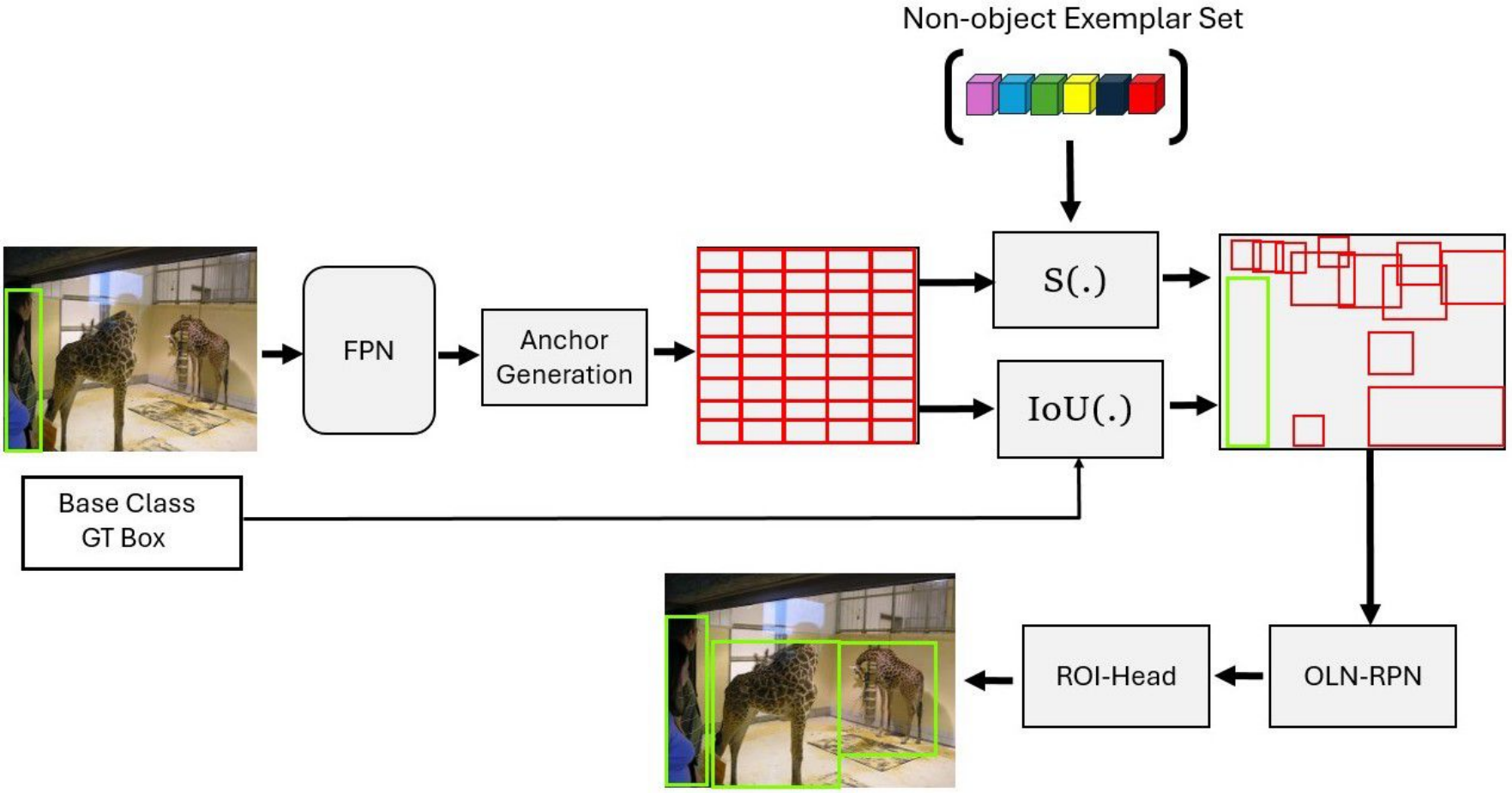}
\caption{Overview of training process with BOWL. First the input image on the left is passed through the feature pyramid network (FPN) backbone.  Next, anchor boxes are generated which cover the image with regions of different positions, aspect ratios and scales.  The anchor boxes are compared with the ground truth known object boxes to generate a set of labeled object boxes (shown in green). With BOWL, a DINO embedding for each image patch within an anchor box is also compared to the non-object exemplar set using cosine similarity, S, and embeddings that have similarity above a certain threshold to an exemplar are used as non-object regions for training. }
\label{fig:model}
\end{figure*}

We incorporate the constructed non-object exemplar set into the training framework of an OWOL model. This integration allows us to explicitly leverage non-object regions during training, leading to more robust object localization.

\paragraph{Network Architecture and Base Training:}
We adopt the object localization network architecture which consists of two primary prediction heads:
\begin{itemize}
   \item A regression head that predicts distances from the feature location to four sides of the ground-truth box $G = (l, \: r, \: t, \: b)$.
   \item An objectness head estimating localization quality $c \in [0,\:1]$ by measuring the alignment between predicted and ground-truth boxes.
\end{itemize}

In the conventional training setup, the objectness head is trained only using foreground proposals that satisfy:
\begin{equation}
   \text{IoU}(A_i, \ B_{gt}) \; \; > \; \; 0.3,
\end{equation}
where $A_i$ is a candidate anchor box and $B_{gt}$ is the set of ground-truth boxes, respectively. The objectness score is supervised using an L1 regression loss:
\begin{equation}
   \cL_{obj} = \|1 - \text{Sigmoid}(c_{A_i})\|_1,
\end{equation}
where $c_{A_i}$ is the centerness score of $A_i$. The value $c_{A_i} = 0$ when $A_i$ has no overlap with the ground-truth box and $c_{A_i} = 1$ when there is perfect overlap. 

\paragraph{Training with Non-object Supervision:}
We observe that the conventional training approach underutilizes non-object regions for supervision. To address this limitation, we augment the training process as shown in Figure \ref{fig:model} by incorporating explicit supervision from our non-object exemplar set $E$. Given the set of exemplar embeddings representing non-object regions, we categorize a region in an image (defined by anchor boxes from the detector) as a non-object region if it is similar to a member of the exemplar set, \ie, the maximum similarity of the image patch embedding to all exemplar patch embeddings is above a fixed threshold. Cosine similarity between patch embeddings is used to measure similarity.  We compute the negative anchor boxes at different scales defined by the detector pipeline. Specifically, for each anchor box $A_i$, we:\\
1. Extract its feature embedding $\phi_{A_i}$ using the same pre-trained model used in constructing $E$. \\
2. Compute its maximum similarity to the exemplar set:
\begin{equation}
   s_{\text{max}}(A_i, \ E) = \max_{e \in E}  \frac {\phi_{P_i} \cdot \phi_{e}}{\left\| \phi_{A_i}\right\| _{2}\left\| \phi_{e}\right\| _{2}}.
\end{equation} \\
3. Label it as a non-object region if:
\begin{equation}
   s_{\text{max}}(A_i, \ E) \; \; > \; \; \gamma,
\end{equation}
where $\gamma$ is a threshold parameter.

For anchor boxes identified as non-object regions, we provide negative supervision to the objectness head by setting their target objectness score to zero. The hypothesis is that the objectness head learns to disregard these regions as the loss will be high relative to ground-truth bounding box. The  objectness loss can thus now be written as:
\begin{equation}
   \cL_{obj}^{new} = \begin{cases}
       \cL_{obj} & \text{, if IoU}(A_i, \ B_{gt}) > 0.3 \\
       1 & \text{, if } s_{\text{min}}(A_i, \ E) \; \; > \; \; \gamma
   \end{cases}.
\end{equation}
$\cL_{obj}^{new}$ ensures that the model learns to distinguish between object and non-object regions more effectively, leading to improved localization performance in open-world settings.

\section{Experiment}
\label{sec:experiment}

\subsection{Experimental Details}

\textbf{Datasets.} We evaluate \ours on two tasks to measure open-set generalization performance: \textbf{(1) Cross-category generalization:} where for a fixed data distribution, the model is evaluated on its performance on object categories not seen during model training, and \textbf{(2) Cross-dataset generalization:} where the model is evaluated on a dataset different from the training dataset and contains unseen object categories. For cross-category evaluation, we use different subsets of the MS-COCO \citep{lin2014microsoft} dataset for training and testing and consider two settings. In the first setting,  we use the subset of MS-COCO containing the 20 object classes from the PASCAL-VOC \citep{everingham2010pascal} for training and utilize the remaining 60 classes of MS-COCO for evaluation. We refer to this setting as VOC to NON-VOC generalization task. For the second setting, we utilize LVIS \cite{gupta2019lvis} benchmark, which provides annotations for 1203 classes in a long-tail distribution. We use the original 80 MS-COCO classes for training and test on the remaining 1123 classes. We refer to this setting as LVIS COCO to LVIS NON COCO generalization task

For cross-dataset evaluation, we train the model on the MS-COCO \citep{lin2014microsoft} dataset containing a mix of object and scene-centric images and evaluate on the ADE20K \citep{zhou2019semantic} dataset (a dataset of scene-centric images). We evaluate our method in two settings: (i) training with COCO subset containing 20 VOC classes and (ii) training with full COCO training set. In both settings, we use ADE20K validation set consisting of instance-level annotations of 3148 object categories.\\
\textbf{Evaluation Metrics.} Following prior work~\citep{kim2021oln, wang2022open}, we calculate Average Recall (AR@$k$) across a range of IoU thresholds, spanning 0.5 to 0.95, and set a default detection budget \( k = 100 \). AR values are presented as percentages. We denote \(\text{AR}_A\) as the AR score for all classes, covering both base and novel categories, while \(\text{AR}_N\) represents the AR score specifically for novel classes. When measuring \(\text{AR}_N\), the detections linked to the base classes are omitted from the budget \( k \). This approach is consistently applied when calculating per-class AR, as well as AR values for objects of different scales: small (\(\text{AR}^s\)), medium (\(\text{AR}^m\)), and large (\(\text{AR}^l\)). \\
\textbf{Implementation Details.} For extracting non-object exemplar patches, we use the pre-trained DINO \citep{caron2021emerging} ViT to extract patch-wise features. Specifically, we extract the last layer (Layer 11) ‘key’ features \citep{amir2021deep} using the \(16\times16\) patch version of ViT S/16 with stride$=8$. Thus each image is represented as a set of $16\times16$ overlapping patches, with an overlap of $8$ pixels. We select $\lambda = 0.2$ for exemplar selection based on held out validation set. From the extracted set of exemplars, we choose $N=1000$ for training \ours. An ablation on the effect of different values for $N$ is included in the supplemental material. 

For open-world localization, we follow the same architecture as OLN \citep{kim2021oln}, which uses the Faster RCNN \citep{ren2015faster} architecture with the ResNet-50 backbone \citep{he2016deep} pretrained on ImageNet \citep{deng2009imagenet}. We implement \ours using the MMDetection framework \citep{chen2019mmdetection} and use the SGD optimizer with an initial learning rate of 0.01. We train \ours for 16 epochs. \ours is trained on 8 NVIDIA 1080Ti GPUs with a batch size of 2 images per device. For non-object anchor box selection, we set $\gamma$ adaptively using Otsu's method \cite{otsu1975threshold}.

\begin{figure*}[t]
\centering
\includegraphics[width=0.92\linewidth]{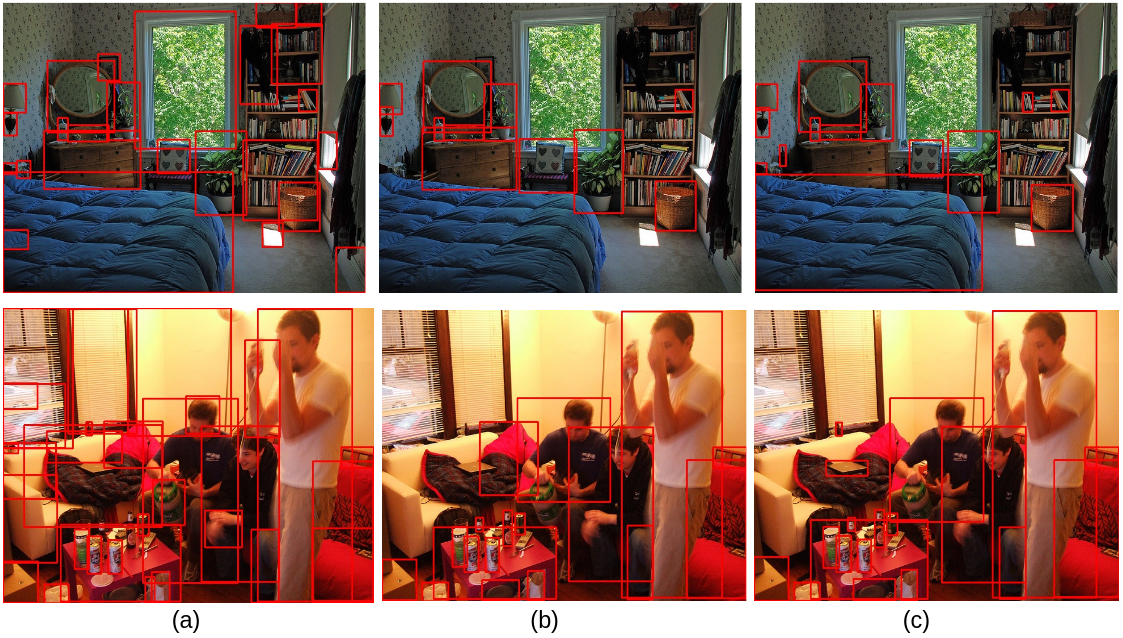}
\caption{\textbf{Qualitative results of (a) GGN \cite{wang2019region}, (b) OLN \cite{kim2021oln}, and (c) \ours on MS-COCO validation images.} The shown results are all predictions with objectness score higher than $0.75$ generated from models trained on VOC categories. From the results we can see that while both OLN and GGN are able to localize unseen novel objects, there are significant false positive and false negative predictions. Specifically, because of noisy pseudo-annotations, GGN incorrectly predicts non-object regions as objects with high objectness score (false predictions on floor and bed in first row while wall and in second row). OLN on the other hand is not able to predict objects with shapes and scales not present in training data due to only using object supervision (false negative prediction of bed in first row and laptop in second row). \ours mitigates the above issues by utilizing non-object supervision leading to better localization of objects. We provide more results in supplementary.}
\label{fig:qualitative}
\end{figure*}

\subsection{Cross-category generalization}

We show the results for cross-category evaluation in Table \ref{table:cross-category} for VOC to COCO classes domain generalization and Table \ref{table:cross-category-lvis} for COCO to NON-COCO domain generalization using annotations from LVIS \cite{gupta2019lvis} benchmark. We report Average Recall score for novel classes (\(\text{AR}_N@100\)) as our primary objective for cross-category evaluation is to assess the model's ability to localize instances from unseen novel categories.

From both results, we see that \ours outperforms other baselines. Specifically, we observe an increase of $4.3\%$ \(\text{AR}_N@100\) compared to OLN for VOC to COCO generalization task. Furthermore, \ours also outperforms OLN  on COCO to LVIS NON-COCO generalization task by $0.9\%$ \(\text{AR}_N@100\). Considering both results, we can conclude that non-object supervision does improve performance for open-world localization task. While background supervision is also used in training the class-agnostic Faster-RCNN baseline, it significantly under performs in comparison to all the other methods. Thus improved results shown by our method can be attributed to accurate selection of non-object regions during model training using our non-object exemplar set. 

\begin{table}[t]
\centering
\caption{Cross-category generalization evaluation \textbf{(VOC $\rightarrow$ COCO)}. }
\resizebox{.9\linewidth}{!}{
\begin{tabular}{ccccc}
\toprule[1pt]
Method & $\text{AR}_N$ & $\text{AR}_{N}^s$ & $\text{AR}_{N}^m$ & $\text{AR}_{N}^l$ \\
\midrule[.75pt]
FRCNN (oracle) & 52.6 & 37.1 & 60.0 & 73.1 \\
\midrule[.5pt]
FRCNN (cls-agn)  &  27.3 & 10.8 & 30.2 & 55.8 \\
OLN \citep{kim2021oln}  & 33.2 & 18.7 & 39.3 & 58.6  \\
GGN \citep{wang2022open} & 31.5 & 11.8 & 37.4 & \textbf{63.8}  \\
\ourso & \textbf{37.5} & \textbf{20.7} & \textbf{46.9} & 58.6 \\
\bottomrule[1pt]
\end{tabular}
}

\label{table:cross-category}
\end{table}

\begin{table}[t]
\centering
\caption{Cross-category generalization evaluation \textbf{(LVIS COCO $\rightarrow$ LVIS NON-COCO)}. }
\resizebox{.9\linewidth}{!}{
\begin{tabular}{ccccc}
\toprule[1pt]
Method & $\text{AR}_N$ & $\text{AR}_{N}^s$ & $\text{AR}_{N}^m$ & $\text{AR}_{N}^l$ \\
\midrule[.75pt]
FRCNN (cls-agn)  &  21.0 & 14.9 & 32.7 & 36.2  \\
OLN \citep{kim2021oln}  & 27.4 & 17.9 & 44.7 & 53.1 \\
GGN \citep{wang2022open} & 22.5 & 15.7 & 35.5 & 38.4  \\
\ourso & \textbf{28.3} & \textbf{18.3} & \textbf{46.5} & \textbf{53.9} \\
\bottomrule[1pt]
\end{tabular}
}

\label{table:cross-category-lvis}
\end{table}


\subsection{Open-set localization in the wild}

\begin{table}[t]
\centering
\caption{Cross-dataset generalization evaluation \textbf{(VOC $\rightarrow$ ADE20K)}.}
\resizebox{.9\linewidth}{!}{
\begin{tabular}{ccccc}
\toprule[1pt]
Method & $\text{AR}_A$ & $\text{AR}_{A}^s$ & $\text{AR}_{A}^m$ & $\text{AR}_{A}^l$ \\
\midrule[.75pt]
FRCNN (cls-agn)  & 22.6 & 15.5 & 23.7 & 26.5  \\
OLN \citep{kim2021oln} & 29.2 & 19.7 & 30.7 & 34.4 \\
GGN \citep{wang2022open} & 27.0 & 16.9 & 27.5 & 33.6  \\
\ourso & \textbf{30.2} &  \textbf{20.7} & \textbf{32.5} &  \textbf{34.8} \\
\bottomrule[1pt]
\end{tabular}
}

\label{table:cross-dataset-voc}
\end{table}

\begin{table}[t]
\centering
\caption{Cross-dataset generalization evaluation \textbf{(COCO $\rightarrow$ ADE20K)}.}
\resizebox{.9\linewidth}{!}{
\begin{tabular}{ccccc}
\toprule[1pt]
Method & $\text{AR}_A$ & $\text{AR}_{A}^s$ & $\text{AR}_{A}^m$ & $\text{AR}_{A}^l$ \\
\midrule[.75pt]
FRCNN (cls-agn)  & 25.9 & 20.5 & 28.5 & 27.4 \\
OLN \citep{kim2021oln} & 32.9 & 25.1 & 35.9 & 35.6\\
GGN \citep{wang2022open} & 29.8 & 18.9 & 29.1 & \textbf{38.2} \\
\ourso & \textbf{34.1} &  \textbf{25.3} & \textbf{37.5} &  37.0 \\
\bottomrule[1pt]
\end{tabular}
}

\label{table:cross-dataset}
\end{table}

A key requirement of any good open-world model is its ability to generalize in out-of-distribution (OOD) settings. Specifically, for the case of open-world object localization, it is important to quantify OOD generalization performance to make sure that the model is not overfitting to a particular dataset. To this end, we evaluate our method for the cross-dataset generalization task, where we train our model on MS-COCO and test it on the ADE20K dataset. For evaluation, we report Average Recall score for all categories (\(\text{AR}_A@100\)). Specifically we report AR score for all categories, since unlike the cross-category setting, similar classes across two datasets can show different properties like change in prototypical shape, scale, size \etc.

We show our results for VOC to ADE20K and COCO to ADE20K in Table \ref{table:cross-dataset-voc} and  \ref{table:cross-dataset}, respectively. Both the results show a  trend, similar to cross-category evaluation tasks, where \ours outperforms other baseline methods. Our method improves over OLN by $1\%$ \(\text{AR}_A@100\) when the model is trained on VOC classes and $1.2\%$ \(\text{AR}_N@100\) when the full COCO dataset is used for training. Improvement in both the settings suggests that our non-object supervision not only improves open-world localization but is also generalizable across data distribution shift. 

\subsection{Ablation Study} 

\textbf{Comparison of different approaches for non-object supervision.} We evaluate our approach for the identification of non-object regions with other methods with respect to improving localization performance. Specifically, we consider gradient energy \cite{avidan2023seam} and image self-correlation map using DINO \cite{caron2021emerging} to identify non-object regions and use the same for training the localization model. Gradient energy has been previously used in figure-ground segmentation methods \cite{yin2009shape, yu2001hierarchical, yu2003object, ren2006figure}, where background is defined as any region in an image with below threshold gradient energy. Similarly, recent works in unsupervised object localization \cite{simeoni2021localizing, simeoni2024unsupervised} utilize self-correlation maps generated using patch features from DINO to identify foreground regions, by considering patches with high correlation to majority patches in an image as background regions.

For both methods, we first generate a binary map specifying foreground and background regions, which is then used to select anchor boxes when training the model. We categorize an anchor box as non-object, if it has at least a 90\% overlap with background regions. 

We compare our approach with the above methods for the VOC to COCO cross-category task. As shown in Table \ref{table:non-object}, all three methods perform better than OLN. This further confirms our initial hypothesis that accurate non-object supervision should lead to improvement in open-world object localization. Amongst the three approaches, our method performs better than the other two methods. We surmise that this is because unlike the other two approaches, our method considers dataset-level statistics to model non-object information. This makes our method more precise in identifying non-object regions. Furthermore, this result also confirms that it is more important to have higher precision in identifying non-object regions for improving open-world object localization.

\begin{table}[t]
\centering
\caption{Cross-category generalization evaluation \textbf{(VOC $\rightarrow$ NON-VOC)} of different sources for non-object supervision.}
\resizebox{.9\linewidth}{!}{
\begin{tabular}{ccccc}
\toprule[1pt]
Method & $\text{AR}_N$ & $\text{AR}_{N}^s$ & $\text{AR}_{N}^m$ & $\text{AR}_{N}^l$ \\
\midrule[.75pt]
Gradient Energy & 36.2 & \textbf{20.9} & 43.1 & 57.4\\
DINO  & 36.8 & 20.3 & 45.4 & 57.7 \\
\ourso & \textbf{37.5} & 20.7 & \textbf{46.9} & \textbf{58.6} \\
\bottomrule[1pt]
\end{tabular}
}

\label{table:non-object}
\end{table}

\begin{figure}[t]
\centering
\includegraphics[width=\linewidth]{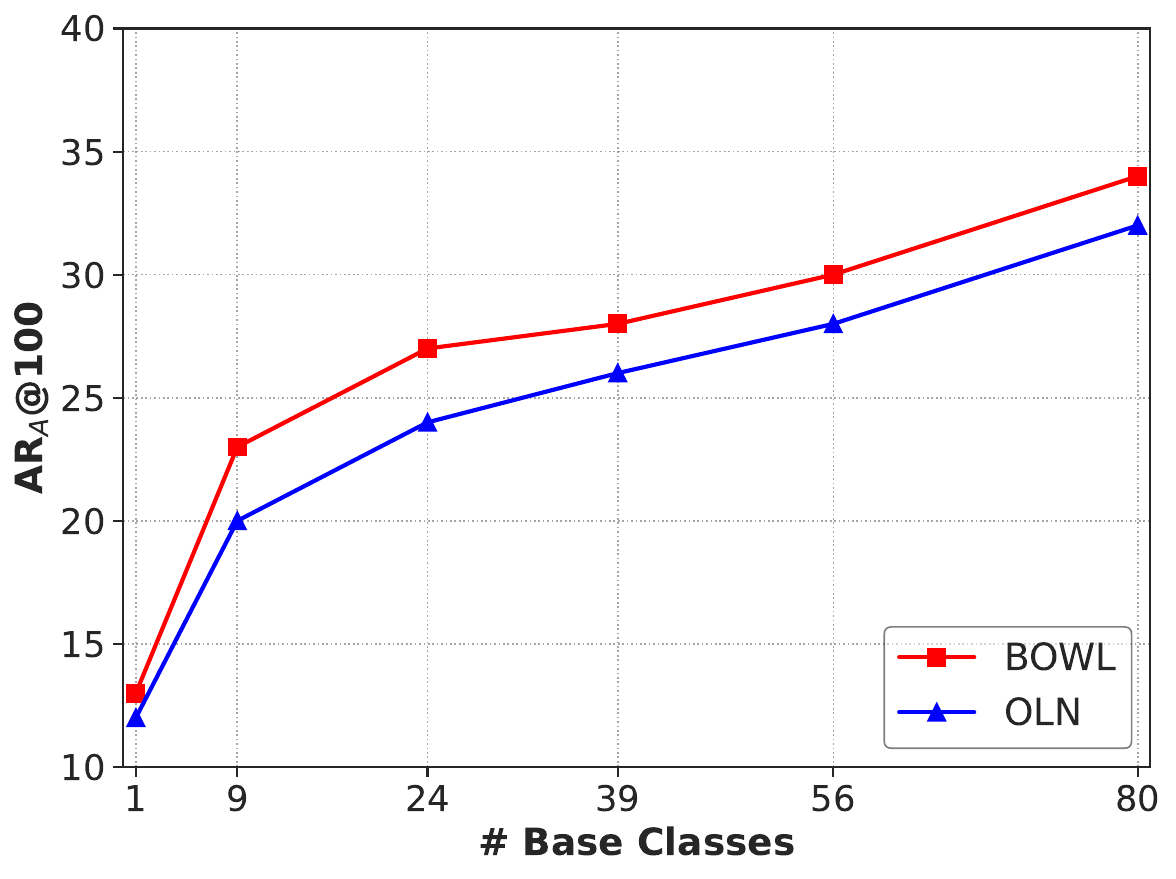}
\caption{Base class evaluation on ADE20K}
\label{fig:ablation_base_class}
\end{figure}

\paragraph{Influence of the number of base classes for training.} One of the advantages of open-world models is that we don't need to annotate every single object instance as these models should be able to generalize to unseen novel objects. Thus it is important to consider the effect of number of base classes used for training the model. To this end, we split the COCO dataset based on the supercategories to create six different splits (Table \ref{table:base-class-ade20k}) and then trained BOWL and OLN on these splits. For evaluation, we test both methods on ADE20K validation set. As shown in Figure \ref{fig:ablation_base_class}, both BOWL and OLN achieve better $\text{AR}_A@100$ scores as the number base classes in training set increases. While for the extreme case of a single base class (`Person') both methods show similar results, we see more improvement in localization performance for \ours when number of classes is on the lower side. Thus \ours is more efficient when considering data annotation cost. 

\begin{table}[t]
\centering
\caption{Base class selection for evaluating the effect of the number of base classes. }
\resizebox{\linewidth}{!}{
\begin{tabular}{@{\hspace{1.5em}}l@{\hspace{1em}}ccc}
\toprule[1pt]
Superclasses &  \texttt{\#}Training Classes & \texttt{\#}Training Images \\
\midrule[.75pt]
\hspace{1.7em}Person & 1 & 64115\\
\textcolor{ForestGreen}{\large\textbf{(+)}} Vehicle & 9 & 74152  \\
\textcolor{ForestGreen}{\large\textbf{(+)}} Outdoor, Animal & 24 & 92169 \\
\textcolor{ForestGreen}{\large\textbf{(+)}} Accessory, Sports & 39 &  93939\\
\textcolor{ForestGreen}{\large\textbf{(+)}} Kitchen, Food & 56 & 107036 \\
\textcolor{ForestGreen}{\large\textbf{(+)}} Furniture, Electronic, & \multirow{2}{*}{80} & \multirow{2}{*}{117266}\\
\hspace{1.7em}Appliance, Indoor & & \\
\bottomrule[1pt]
\end{tabular}
}

\label{table:base-class-ade20k}
\end{table}

\section{Discussion and Conclusion}
\label{sec:conclusion}

In this work, we introduce the idea of learning a model of non-object regions in a self-supervised way and use it to improve the accuracy of open-world object localization.  We have shown that this idea improves results on standard datasets compared to the state-of-the-art. \ours is the first method that uses a generic model of non-object regions to improve open-world object localization.  Our ablation experiments confirm that our exemplar-based non-object model achieves better results than simpler background models, while also showing that our improvement over the OLN open-world detector is independent of the number of base classes in the training set.  For future work, we think further gains can be realized by combining the idea of background modeling with the creation of pseudo-labels to increase the number of labeled object for training.

{\bf Limitations:} While incorporating a background model into open-world localization training improves accuracy, it does require the extra computational expense of learning a background model, although this only needs to be done once. 
Also, \ours is slightly worse, in terms of performance, on large objects than some competing methods.  This is likely due to competing methods' use of pseudo annotations which add more large object training examples.
{
    \small
    \bibliographystyle{ieeenat_fullname}
    \bibliography{main}

\begin{thebibliography}{41}
\providecommand{\natexlab}[1]{#1}
\providecommand{\url}[1]{\texttt{#1}}
\expandafter\ifx\csname urlstyle\endcsname\relax
  \providecommand{\doi}[1]{doi: #1}\else
  \providecommand{\doi}{doi: \begingroup \urlstyle{rm}\Url}\fi

\bibitem[Alexe et~al.(2012)Alexe, Deselaers, and Ferrari]{alexe2012measuring}
Bogdan Alexe, Thomas Deselaers, and Vittorio Ferrari.
\newblock Measuring the objectness of image windows.
\newblock \emph{IEEE transactions on pattern analysis and machine intelligence}, 34\penalty0 (11):\penalty0 2189--2202, 2012.

\bibitem[Amir et~al.()Amir, Gandelsman, Bagon, and Dekel]{amir2021deep}
Shir Amir, Yossi Gandelsman, Shai Bagon, and Tali Dekel.
\newblock Deep vit features as dense visual descriptors.

\bibitem[Arbel{\'a}ez et~al.(2014)Arbel{\'a}ez, Pont-Tuset, Barron, Marques, and Malik]{arbelaez2014multiscale}
Pablo Arbel{\'a}ez, Jordi Pont-Tuset, Jonathan~T Barron, Ferran Marques, and Jitendra Malik.
\newblock Multiscale combinatorial grouping.
\newblock In \emph{Proceedings of the IEEE conference on computer vision and pattern recognition}, pages 328--335, 2014.

\bibitem[Avidan and Shamir(2023)]{avidan2023seam}
Shai Avidan and Ariel Shamir.
\newblock Seam carving for content-aware image resizing.
\newblock In \emph{Seminal Graphics Papers: Pushing the Boundaries, Volume 2}, pages 609--617. 2023.

\bibitem[Borji et~al.(2019)Borji, Cheng, Hou, Jiang, and Li]{borji2019salient}
Ali Borji, Ming-Ming Cheng, Qibin Hou, Huaizu Jiang, and Jia Li.
\newblock Salient object detection: A survey.
\newblock \emph{Computational visual media}, 5:\penalty0 117--150, 2019.

\bibitem[Caron et~al.(2021)Caron, Touvron, Misra, J{\'e}gou, Mairal, Bojanowski, and Joulin]{caron2021emerging}
Mathilde Caron, Hugo Touvron, Ishan Misra, Herv{\'e} J{\'e}gou, Julien Mairal, Piotr Bojanowski, and Armand Joulin.
\newblock Emerging properties in self-supervised vision transformers.
\newblock In \emph{Proceedings of the IEEE/CVF international conference on computer vision}, pages 9650--9660, 2021.

\bibitem[Chen et~al.(2019)Chen, Wang, Pang, Cao, Xiong, Li, Sun, Feng, Liu, Xu, et~al.]{chen2019mmdetection}
Kai Chen, Jiaqi Wang, Jiangmiao Pang, Yuhang Cao, Yu Xiong, Xiaoxiao Li, Shuyang Sun, Wansen Feng, Ziwei Liu, Jiarui Xu, et~al.
\newblock Mmdetection: Open mmlab detection toolbox and benchmark.
\newblock \emph{arXiv preprint arXiv:1906.07155}, 2019.

\bibitem[Deng et~al.(2009)Deng, Dong, Socher, Li, Li, and Fei-Fei]{deng2009imagenet}
Jia Deng, Wei Dong, Richard Socher, Li-Jia Li, Kai Li, and Li Fei-Fei.
\newblock Imagenet: A large-scale hierarchical image database.
\newblock In \emph{2009 IEEE conference on computer vision and pattern recognition}, pages 248--255. Ieee, 2009.

\bibitem[Endres and Hoiem(2010)]{endres2010category}
Ian Endres and Derek Hoiem.
\newblock Category independent object proposals.
\newblock In \emph{Computer Vision--ECCV 2010: 11th European Conference on Computer Vision, Heraklion, Crete, Greece, September 5-11, 2010, Proceedings, Part V 11}, pages 575--588. Springer, 2010.

\bibitem[Everingham et~al.(2010)Everingham, Van~Gool, Williams, Winn, and Zisserman]{everingham2010pascal}
Mark Everingham, Luc Van~Gool, Christopher~KI Williams, John Winn, and Andrew Zisserman.
\newblock The pascal visual object classes (voc) challenge.
\newblock \emph{International journal of computer vision}, 88:\penalty0 303--338, 2010.

\bibitem[Gupta et~al.(2019)Gupta, Dollar, and Girshick]{gupta2019lvis}
Agrim Gupta, Piotr Dollar, and Ross Girshick.
\newblock Lvis: A dataset for large vocabulary instance segmentation.
\newblock In \emph{Proceedings of the IEEE/CVF conference on computer vision and pattern recognition}, pages 5356--5364, 2019.

\bibitem[He et~al.(2016)He, Zhang, Ren, and Sun]{he2016deep}
Kaiming He, Xiangyu Zhang, Shaoqing Ren, and Jian Sun.
\newblock Deep residual learning for image recognition.
\newblock In \emph{Proceedings of the IEEE conference on computer vision and pattern recognition}, pages 770--778, 2016.

\bibitem[Jones and Rehg(2002)]{jones2002statistical}
Michael~J Jones and James~M Rehg.
\newblock Statistical color models with application to skin detection.
\newblock \emph{International journal of computer vision}, 46:\penalty0 81--96, 2002.

\bibitem[Kim et~al.(2022)Kim, Lin, Angelova, Kweon, and Kuo]{kim2021oln}
Dahun Kim, Tsung-Yi Lin, Anelia Angelova, In~So Kweon, and Weicheng Kuo.
\newblock Learning open-world object proposals without learning to classify.
\newblock \emph{IEEE Robotics and Automation Letters (RA-L)}, 2022.

\bibitem[Kuo et~al.(2015)Kuo, Hariharan, and Malik]{kuo2015deepbox}
Weicheng Kuo, Bharath Hariharan, and Jitendra Malik.
\newblock Deepbox: Learning objectness with convolutional networks.
\newblock In \emph{Proceedings of the IEEE international conference on computer vision}, pages 2479--2487, 2015.

\bibitem[Li et~al.(2019)Li, Liu, Ouyang, and Wang]{li2019zoom}
Hongyang Li, Yu Liu, Wanli Ouyang, and Xiaogang Wang.
\newblock Zoom out-and-in network with map attention decision for region proposal and object detection.
\newblock \emph{International Journal of Computer Vision}, 127:\penalty0 225--238, 2019.

\bibitem[Lin et~al.(2014)Lin, Maire, Belongie, Hays, Perona, Ramanan, Doll{\'a}r, and Zitnick]{lin2014microsoft}
Tsung-Yi Lin, Michael Maire, Serge Belongie, James Hays, Pietro Perona, Deva Ramanan, Piotr Doll{\'a}r, and C~Lawrence Zitnick.
\newblock Microsoft coco: Common objects in context.
\newblock In \emph{Computer Vision--ECCV 2014: 13th European Conference, Zurich, Switzerland, September 6-12, 2014, Proceedings, Part V 13}, pages 740--755. Springer, 2014.

\bibitem[Manen et~al.(2013)Manen, Guillaumin, and Van~Gool]{manen2013prime}
Santiago Manen, Matthieu Guillaumin, and Luc Van~Gool.
\newblock Prime object proposals with randomized prim's algorithm.
\newblock In \emph{Proceedings of the IEEE international conference on computer vision}, pages 2536--2543, 2013.

\bibitem[O~Pinheiro et~al.(2015)O~Pinheiro, Collobert, and Doll{\'a}r]{o2015learning}
Pedro~O O~Pinheiro, Ronan Collobert, and Piotr Doll{\'a}r.
\newblock Learning to segment object candidates.
\newblock \emph{Advances in neural information processing systems}, 28, 2015.

\bibitem[Oquab et~al.(2023)Oquab, Darcet, Moutakanni, Vo, Szafraniec, Khalidov, Fernandez, Haziza, Massa, El-Nouby, et~al.]{oquab2023dinov2}
Maxime Oquab, Timoth{\'e}e Darcet, Th{\'e}o Moutakanni, Huy Vo, Marc Szafraniec, Vasil Khalidov, Pierre Fernandez, Daniel Haziza, Francisco Massa, Alaaeldin El-Nouby, et~al.
\newblock Dinov2: Learning robust visual features without supervision.
\newblock \emph{arXiv preprint arXiv:2304.07193}, 2023.

\bibitem[Otsu et~al.(1975)]{otsu1975threshold}
Nobuyuki Otsu et~al.
\newblock A threshold selection method from gray-level histograms.
\newblock \emph{Automatica}, 11\penalty0 (285-296):\penalty0 23--27, 1975.

\bibitem[Ren(2015)]{ren2015faster}
Shaoqing Ren.
\newblock Faster r-cnn: Towards real-time object detection with region proposal networks.
\newblock \emph{arXiv preprint arXiv:1506.01497}, 2015.

\bibitem[Ren et~al.(2006)Ren, Fowlkes, and Malik]{ren2006figure}
Xiaofeng Ren, Charless~C Fowlkes, and Jitendra Malik.
\newblock Figure/ground assignment in natural images.
\newblock In \emph{Computer Vision--ECCV 2006: 9th European Conference on Computer Vision, Graz, Austria, May 7-13, 2006. Proceedings, Part II 9}, pages 614--627. Springer, 2006.

\bibitem[Rezatofighi et~al.(2019)Rezatofighi, Tsoi, Gwak, Sadeghian, Reid, and Savarese]{rezatofighi2019generalized}
Hamid Rezatofighi, Nathan Tsoi, JunYoung Gwak, Amir Sadeghian, Ian Reid, and Silvio Savarese.
\newblock Generalized intersection over union: A metric and a loss for bounding box regression.
\newblock In \emph{Proceedings of the IEEE/CVF conference on computer vision and pattern recognition}, pages 658--666, 2019.

\bibitem[Saito et~al.(2022)Saito, Hu, Darrell, and Saenko]{saito2022learning}
Kuniaki Saito, Ping Hu, Trevor Darrell, and Kate Saenko.
\newblock Learning to detect every thing in an open world.
\newblock In \emph{European Conference on Computer Vision}, pages 268--284. Springer, 2022.

\bibitem[Sim{\'e}oni et~al.(2021)Sim{\'e}oni, Puy, Vo, Roburin, Gidaris, Bursuc, P{\'e}rez, Marlet, and Ponce]{simeoni2021localizing}
Oriane Sim{\'e}oni, Gilles Puy, Huy~V Vo, Simon Roburin, Spyros Gidaris, Andrei Bursuc, Patrick P{\'e}rez, Renaud Marlet, and Jean Ponce.
\newblock Localizing objects with self-supervised transformers and no labels.
\newblock In \emph{BMVC 2021-32nd British Machine Vision Conference}, 2021.

\bibitem[Sim{\'e}oni et~al.(2023)Sim{\'e}oni, Sekkat, Puy, Vobeck{\`y}, Zablocki, and P{\'e}rez]{simeoni2023unsupervised}
Oriane Sim{\'e}oni, Chlo{\'e} Sekkat, Gilles Puy, Anton{\'\i}n Vobeck{\`y}, {\'E}loi Zablocki, and Patrick P{\'e}rez.
\newblock Unsupervised object localization: Observing the background to discover objects.
\newblock In \emph{Proceedings of the IEEE/CVF Conference on Computer Vision and Pattern Recognition}, pages 3176--3186, 2023.

\bibitem[Sim{\'e}oni et~al.(2024)Sim{\'e}oni, Zablocki, Gidaris, Puy, and P{\'e}rez]{simeoni2024unsupervised}
Oriane Sim{\'e}oni, {\'E}loi Zablocki, Spyros Gidaris, Gilles Puy, and Patrick P{\'e}rez.
\newblock Unsupervised object localization in the era of self-supervised vits: A survey.
\newblock \emph{International Journal of Computer Vision}, pages 1--28, 2024.

\bibitem[Singh et~al.(2012)Singh, Gupta, and Efros]{singh2012unsupervised}
Saurabh Singh, Abhinav Gupta, and Alexei~A Efros.
\newblock Unsupervised discovery of mid-level discriminative patches.
\newblock In \emph{Computer Vision--ECCV 2012: 12th European Conference on Computer Vision, Florence, Italy, October 7-13, 2012, Proceedings, Part II 12}, pages 73--86. Springer, 2012.

\bibitem[Tian et~al.(2020)Tian, Shen, Chen, and He]{tian2020fcos}
Zhi Tian, Chunhua Shen, Hao Chen, and Tong He.
\newblock Fcos: A simple and strong anchor-free object detector.
\newblock \emph{IEEE transactions on pattern analysis and machine intelligence}, 44\penalty0 (4):\penalty0 1922--1933, 2020.

\bibitem[Uijlings et~al.(2013)Uijlings, Van De~Sande, Gevers, and Smeulders]{uijlings2013selective}
Jasper~RR Uijlings, Koen~EA Van De~Sande, Theo Gevers, and Arnold~WM Smeulders.
\newblock Selective search for object recognition.
\newblock \emph{International journal of computer vision}, 104:\penalty0 154--171, 2013.

\bibitem[Von~Luxburg(2007)]{von2007tutorial}
Ulrike Von~Luxburg.
\newblock A tutorial on spectral clustering.
\newblock \emph{Statistics and computing}, 17:\penalty0 395--416, 2007.

\bibitem[Wang et~al.(2019)Wang, Chen, Yang, Loy, and Lin]{wang2019region}
Jiaqi Wang, Kai Chen, Shuo Yang, Chen~Change Loy, and Dahua Lin.
\newblock Region proposal by guided anchoring.
\newblock In \emph{Proceedings of the IEEE/CVF conference on computer vision and pattern recognition}, pages 2965--2974, 2019.

\bibitem[Wang et~al.(2022)Wang, Feiszli, Wang, Malik, and Tran]{wang2022open}
Weiyao Wang, Matt Feiszli, Heng Wang, Jitendra Malik, and Du Tran.
\newblock Open-world instance segmentation: Exploiting pseudo ground truth from learned pairwise affinity.
\newblock In \emph{Proceedings of the IEEE/CVF conference on computer vision and pattern recognition}, pages 4422--4432, 2022.

\bibitem[Wang et~al.(2023)Wang, Shen, Yuan, Du, Li, Hu, Crowley, and Vaufreydaz]{wang2023tokencut}
Yangtao Wang, Xi Shen, Yuan Yuan, Yuming Du, Maomao Li, Shell~Xu Hu, James~L Crowley, and Dominique Vaufreydaz.
\newblock Tokencut: Segmenting objects in images and videos with self-supervised transformer and normalized cut.
\newblock \emph{IEEE transactions on pattern analysis and machine intelligence}, 2023.

\bibitem[Wong and Leow(2000)]{wong2000color}
Swee-Seong Wong and Wee~Kheng Leow.
\newblock Color segmentation and figure-ground segregation of natural images.
\newblock In \emph{Proceedings 2000 International Conference on Image Processing (Cat. No. 00CH37101)}, pages 120--123. IEEE, 2000.

\bibitem[Yin and Collins(2009)]{yin2009shape}
Zhaozheng Yin and Robert~T Collins.
\newblock Shape constrained figure-ground segmentation and tracking.
\newblock In \emph{2009 IEEE Conference on Computer Vision and Pattern Recognition}, pages 731--738. IEEE, 2009.

\bibitem[Yu and Shi(2003)]{yu2003object}
SU Yu and Jianbo Shi.
\newblock Object-specific figure-ground segregation.
\newblock In \emph{2003 IEEE Computer Society Conference on Computer Vision and Pattern Recognition, 2003. Proceedings.}, pages II--39. IEEE, 2003.

\bibitem[Yu et~al.(2001)Yu, Lee, and Kanade]{yu2001hierarchical}
Stella~X Yu, Tai~Sing Lee, and Takeo Kanade.
\newblock A hierarchical markov random field model for figure-ground segregation.
\newblock In \emph{International Workshop on Energy Minimization Methods in Computer Vision and Pattern Recognition}, pages 118--133. Springer, 2001.

\bibitem[Zhou et~al.(2019)Zhou, Zhao, Puig, Xiao, Fidler, Barriuso, and Torralba]{zhou2019semantic}
Bolei Zhou, Hang Zhao, Xavier Puig, Tete Xiao, Sanja Fidler, Adela Barriuso, and Antonio Torralba.
\newblock Semantic understanding of scenes through the ade20k dataset.
\newblock \emph{International Journal of Computer Vision}, 127:\penalty0 302--321, 2019.

\bibitem[Zitnick and Doll{\'a}r(2014)]{zitnick2014edge}
C~Lawrence Zitnick and Piotr Doll{\'a}r.
\newblock Edge boxes: Locating object proposals from edges.
\newblock In \emph{Computer Vision--ECCV 2014: 13th European Conference, Zurich, Switzerland, September 6-12, 2014, Proceedings, Part V 13}, pages 391--405. Springer, 2014.

\end{thebibliography}
}
\section{Supplemental Material}

\begin{figure*}[ht]
    \centering
    \includegraphics[width=0.5\textwidth]{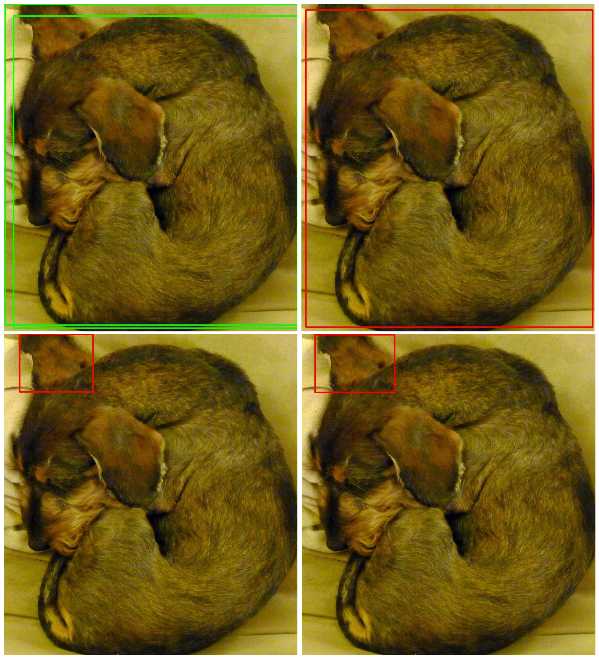}%
     \hspace{0.01\textwidth}%
    \includegraphics[width=0.37\textwidth]{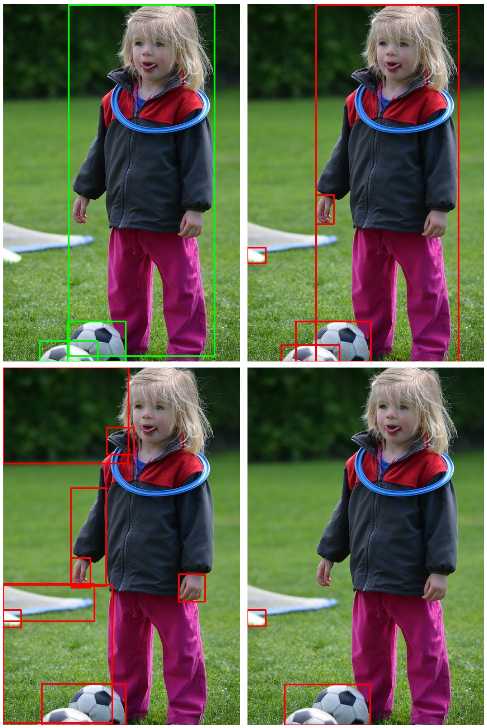}
    \caption{\textbf{Qualitative results of GGN \cite{wang2019region}, OLN \cite{kim2021oln}, and \ours on MS-COCO validation images.} Here For each image, (Row 1, Column 1) $\rightarrow$ Ground Truth bounding boxes; (Row 1, Column 2) $\rightarrow$ \ours predictions; (Row 2, Column 1) $\rightarrow$ GGN predictions; (Row 2, Column 2) $\rightarrow$ OLN predictions. Green colored boxes refer to ground-truth bounding boxes, while red colored boxes are model predictions with objectness score greater than $0.75$ }
    \label{fig:coco_val_1}
\end{figure*}

In this supplemental material, we first provide more qualitative results comparing \ours with competing baseline methods: OLN \cite{kim2021oln} and GGN \cite{wang2022open}. We also provide visualization of the patches selected as exemplars to represent non-object information. Finally we evaluate the accuracy of our exemplar set in identifying non-object regions in unseen images and show visualizations of negative anchor boxes used for training \ours.

\subsection{ More qualitative results}

\begin{figure*}[htbp]
    \centering
    \begin{subfigure}[b]{\textwidth}
        \centering
        \includegraphics[width=0.85\textwidth]{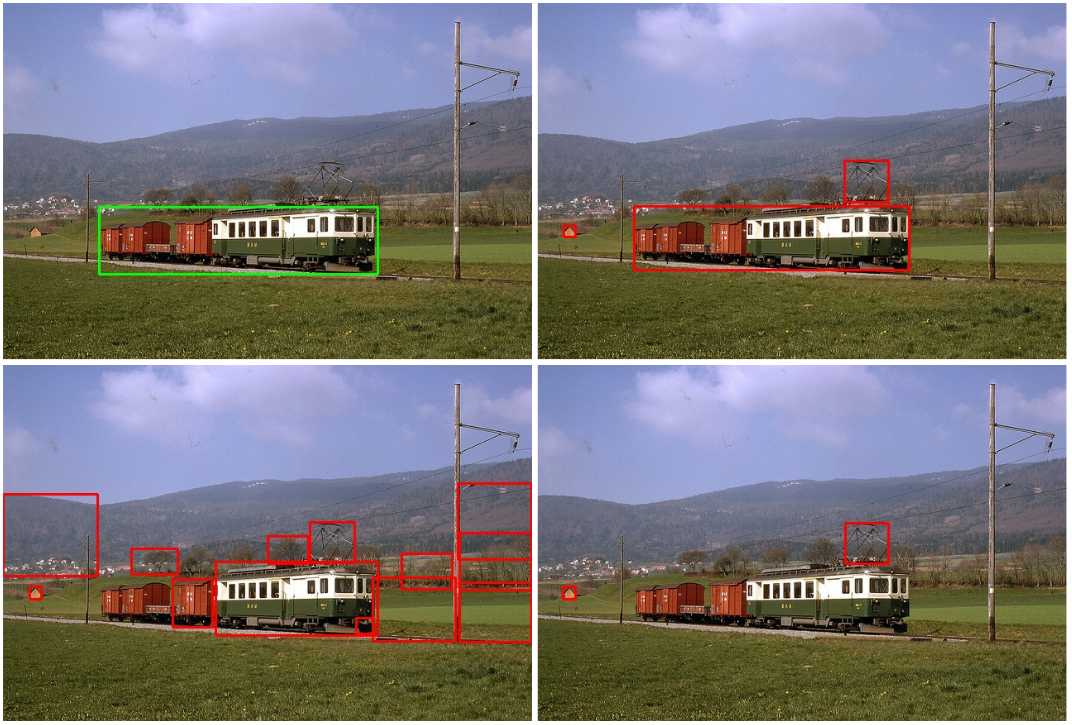}
        \label{fig:fig1}
    \end{subfigure}

    \vspace{0.1em}  
    
    \begin{subfigure}[b]{\textwidth}
        \centering
        \includegraphics[width=0.85\textwidth]{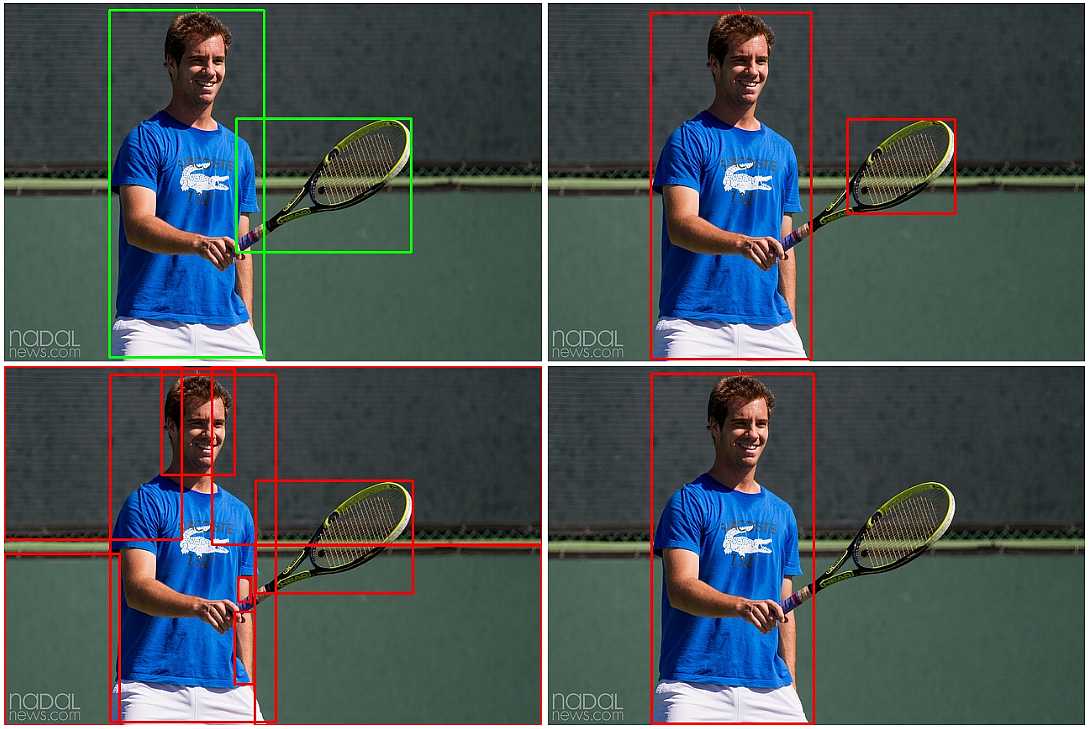}
        \label{fig:fig3}
    \end{subfigure}
     \caption{\textbf{Qualitative results of GGN \cite{wang2019region}, OLN \cite{kim2021oln}, and \ours on MS-COCO validation images.} Here For each image, (Row 1, Column 1) $\rightarrow$ Ground Truth bounding boxes; (Row 1, Column 2) $\rightarrow$ \ours predictions; (Row 2, Column 1) $\rightarrow$ GGN predictions; (Row 2, Column 2) $\rightarrow$ OLN predictions. Green colored boxes refer to ground-truth bounding boxes, while red colored boxes are model predictions with objectness score greater than $0.75$ }
    \label{fig:coco_val_2}
\end{figure*}

\begin{figure*}[htbp]
    \centering
    \begin{subfigure}[b]{\textwidth}
        \centering
        \includegraphics[width=0.85\textwidth]{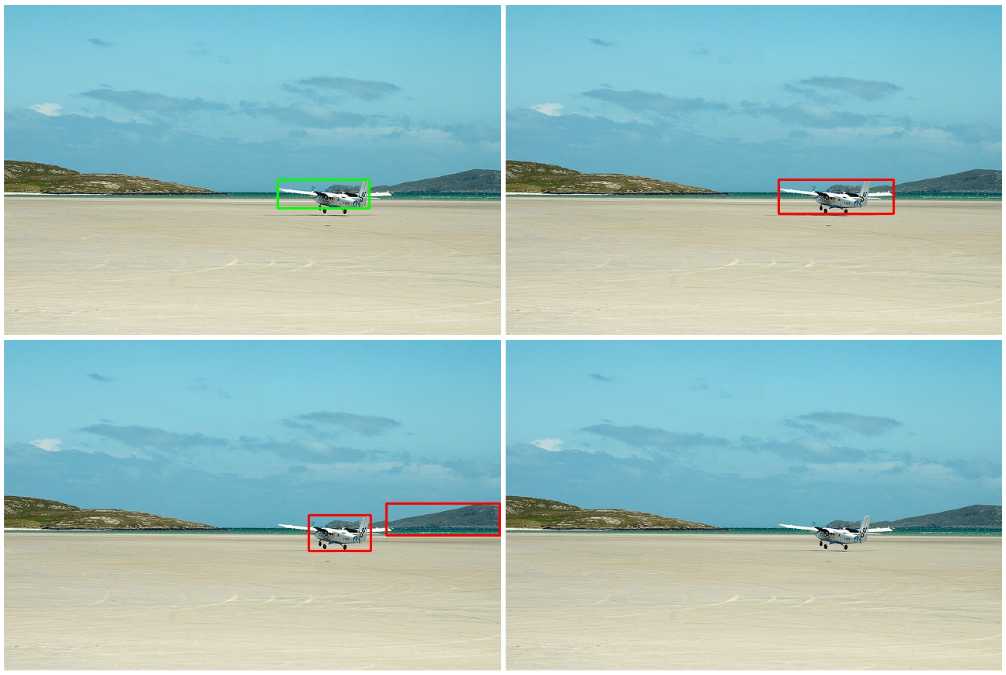}
        \label{fig:fig1b}
    \end{subfigure}

    \vspace{0.1em}  
    
    \begin{subfigure}[b]{\textwidth}
        \centering
        \includegraphics[width=0.85\textwidth]{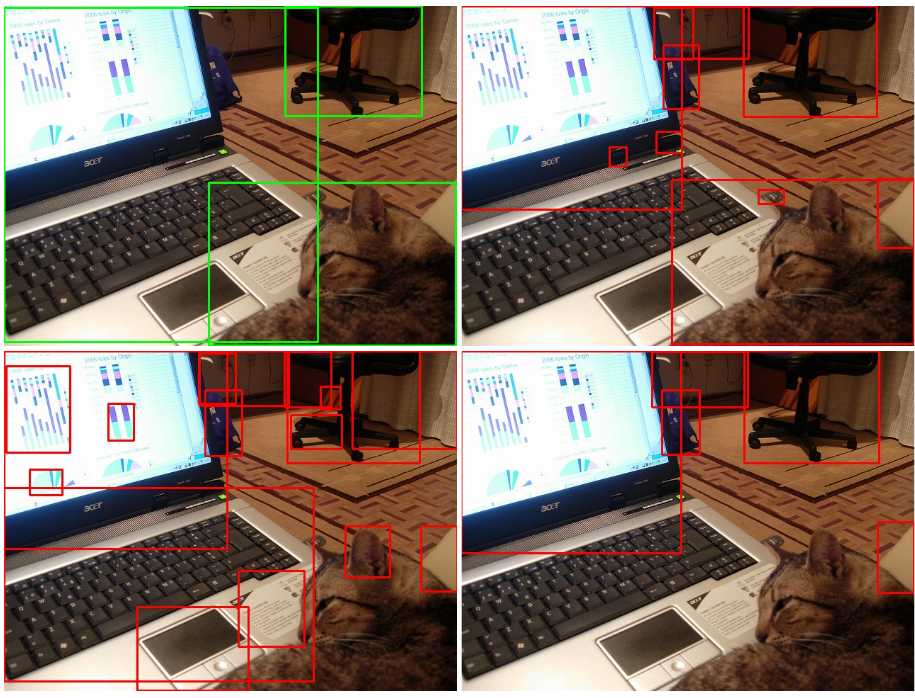}
        \label{fig:fig3b}
    \end{subfigure}
     \caption{\textbf{Qualitative results of GGN \cite{wang2019region}, OLN \cite{kim2021oln}, and \ours on MS-COCO validation images.} Here For each image, (Row 1, Column 1) $\rightarrow$ Ground Truth bounding boxes; (Row 1, Column 2) $\rightarrow$ \ours predictions; (Row 2, Column 1) $\rightarrow$ GGN predictions; (Row 2, Column 2) $\rightarrow$ OLN predictions. Green colored boxes refer to ground-truth bounding boxes, while red colored boxes are model predictions with objectness score greater than $0.75$ }
    \label{fig:coco_val_3}
\end{figure*}

We show localization results of \ours, OLN \cite{kim2021oln} and GGN \cite{wang2022open} on some of the MS-COCO \cite{lin2014microsoft} validation set image in Figures \ref{fig:coco_val_1}, \ref{fig:coco_val_2} and \ref{fig:coco_val_3}. The results are generated using models trained on the $20$ VOC categories. For each method, we show all the predicted boxes with objectness score greater than $0.75$. From all the qualitative results, we observe that \ours provides significantly better results. Specifically, as discussed in the main paper, we see that while both OLN and GGN are able to localize unseen objects, both methods suffer from high false-positive (for GGN) and false-negative predictions (for OLN). These qualitative results further support our hypothesis that non-object supervision can boost objectness learning and improve open-set object localization.

\subsection{Non-object exemplar set}
\begin{figure*}[h]
\centering
\includegraphics[width=1.05\linewidth]{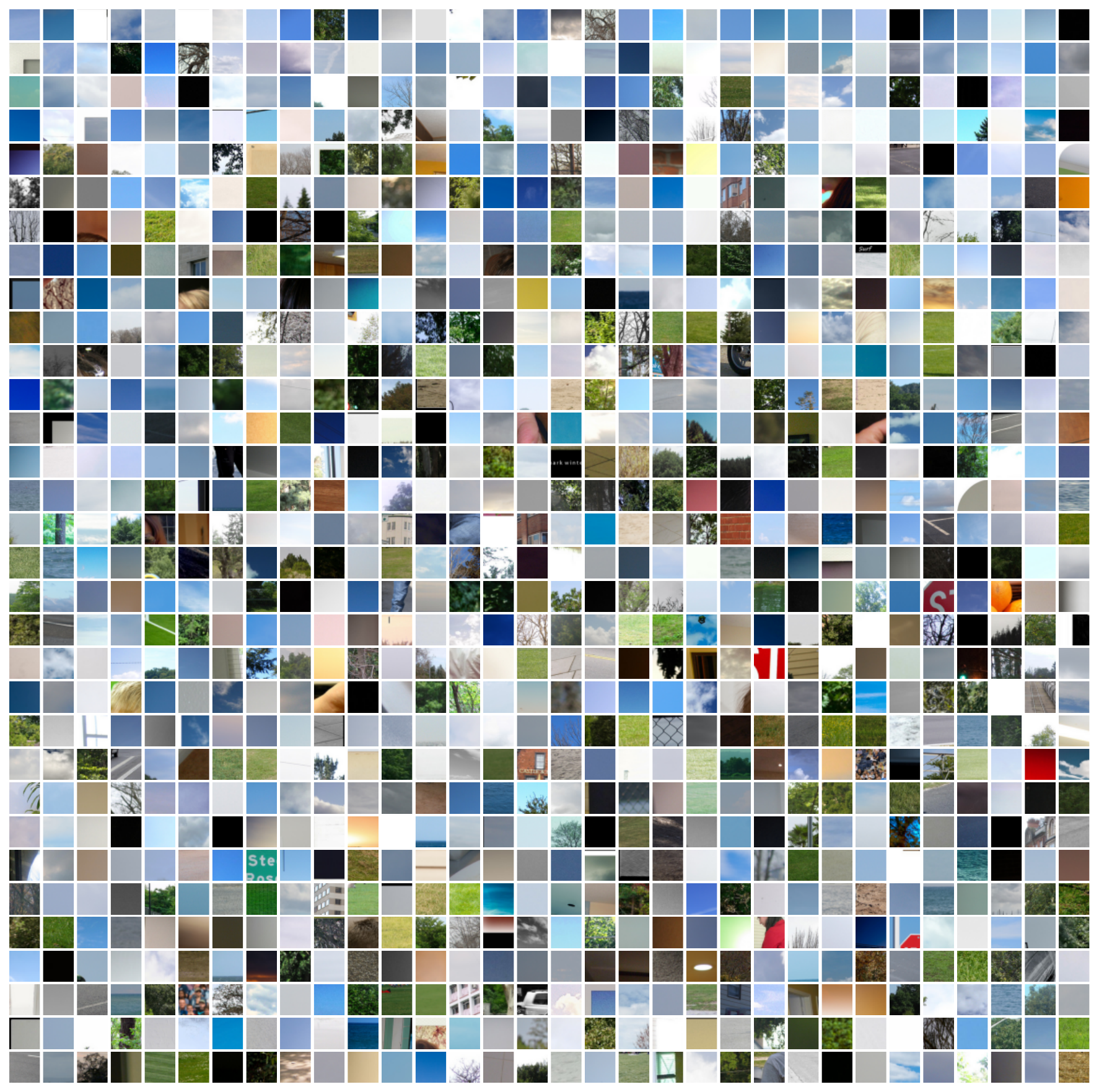}
\caption{Non-object exemplar patches, constructed from the training set of the COCO dataset}
\label{fig:exemplars2}
\end{figure*}

Figure \ref{fig:exemplars2} shows all the non-object exemplar patches used to identify negative anchor boxes when training \ours. Specifically, $86436$ patches were selected in total using our exemplar selection method. Out of the total set, we further selected the top $1000$ patches based on the nearest-neighbor count of each exemplar patch \ie how many patches in the total set of all patches are similar to a given exemplar patch, to create non-object exemplar set. In Figure  \ref{fig:exemplars2}, we show 
these $1000$ non-object patches, in the descending order of their nearest-neighbor count, with patches in the first row representing the most common patches found in MS-COCO \cite{lin2014microsoft} training set, which were used to represent the total set of patches.
As can be observed from Figure \ref{fig:exemplars2}, all the patches represent regions generally characterized as background, for example sky, forest \etc. Furthermore, the order of the exemplar patches also aligns with the frequency of the background semantic regions in the dataset, for example, sky is more common than grassy regions in the dataset. We can also see that the selected exemplar set is visually and semantically diverse, leading to a compact model of non-objectness in the dataset.

\subsection{Ablation study: Number of exemplar samples to model non-objectness}

\begin{table}[h]
\centering
\caption{Our model's performance with varying sizes of the exemplar set (constructed from the MS-COCO training set) for identifying non-object regions on MS-COCO validation set.}
\resizebox{.9\linewidth}{!}{
\begin{tabular}{ccc}
\toprule[1pt]
Num Exemplars & Percentage & Average precision (\%) \\
\midrule[.75pt]
N = 432 & 0.5\% & 95.23 \\
N = 864  & 1\% & 95.10 \\
N = 1000  & 1.15\% & 95.08\\
N = 4321  & 5\% & 94.57 \\
N =  8643 & 10\% & 94.50 \\
N = 21609 & 25\% & 94.48 \\
\bottomrule[1pt]
\end{tabular}
}
\label{table:num-exemplar}
\end{table}

As mentioned in the main paper and in the previous section, after extracting the exemplar set of patches, we further subsample exemplars based on the number of nearest-neighbor counts of exemplars. Specifically we select the top $N$ exemplar samples that are most similar to other patches in the original set of all patches. For training \ours, we selected $N = 1000$ exemplar samples to create a non-object exemplar subset representing a compact model of non-objectness. To further validate our design choice, we conduct an experiment to measure the precision of non-object regions identified in unseen images by varying the size of the non-object exemplar set. Concretely, for a given set of non-object exemplar patch embeddings, we categorize patches in a test image as object or non-object by computing the similarity of the test image patch with the exemplar set. Given the binary segmentation of the test image, we compute anchor boxes of a fixed resolution that overlaps with non-object regions. We then compute the overlap of the non-object anchor boxes with ground-truth object boxes from all classes. Based on the overlap between predicted non-object anchor box and ground-truth object bounding box, we calculate the precision of the non-object anchor boxes. The above setup simulates our process of identifying negative anchor boxes used for training \ours. For the above experiment, we measure the overlap with IoU threshold of $0.1$ and fix the anchor box size to $128 \times 128$. We conduct the above experiment on MS-COCO \cite{lin2014microsoft}  validation set with ground-truth bounding boxes from  all the 80 COCO categories. We report our results in Table \ref{table:num-exemplar}. From the results, we can see that with a smaller subset we obtain the highest average precision. As we increase the size of the subset, the precision reduces by a small margin but saturates quickly. This result validates our design choice and confirms our hypothesis that a small subset of most common exemplar patches is sufficient to accurately identify non-object regions in unseen images.

\subsection{Negative anchor box visualizations}
\begin{figure*}[h]
    \centering
    \begin{subfigure}[b]{\textwidth}
        \centering
        \includegraphics[width=0.8\textwidth]{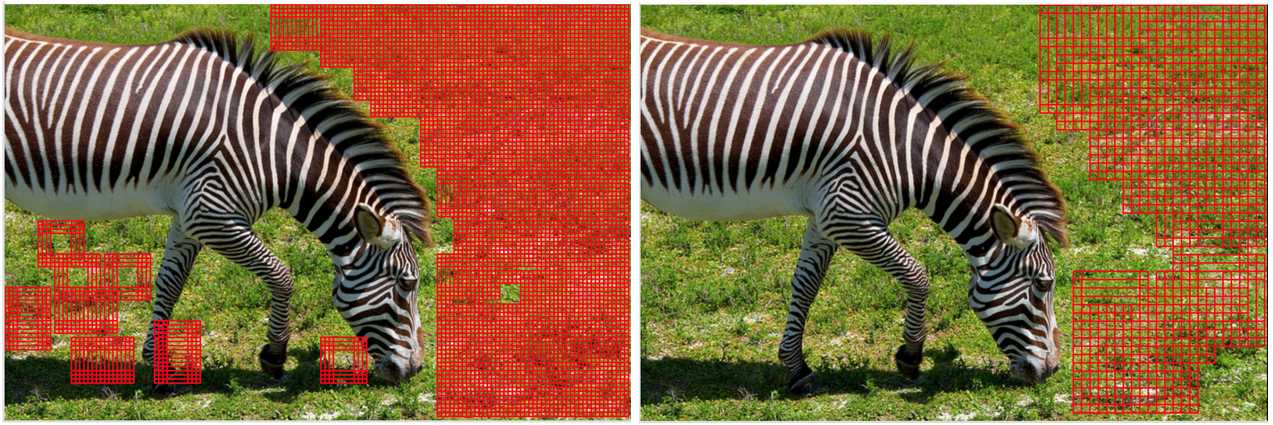}
        \label{fig:fig1}
    \end{subfigure}

    \vspace{0.1em}  
    
    \begin{subfigure}[b]{\textwidth}
        \centering
        \includegraphics[width=0.8\textwidth]{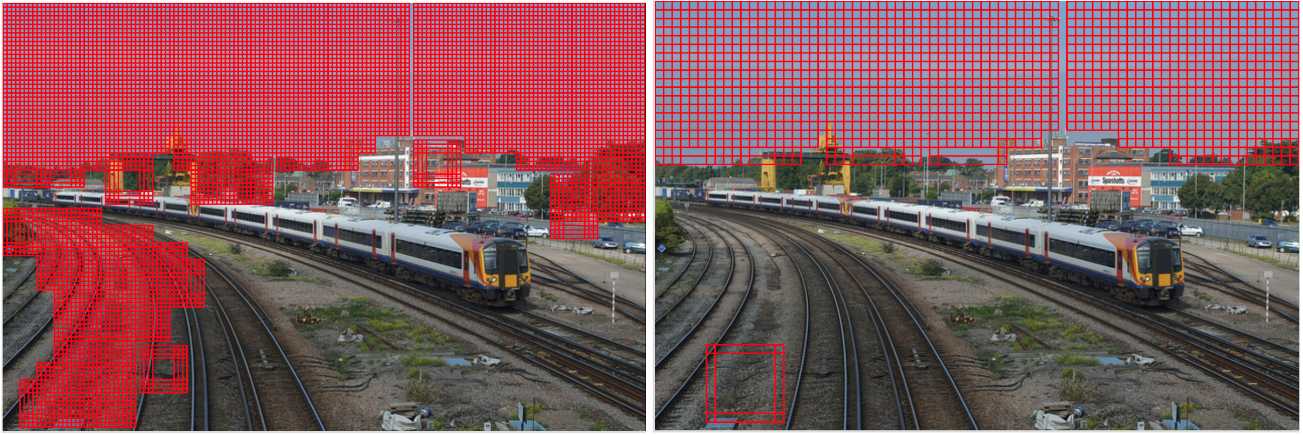}
        \label{fig:fig3}
    \end{subfigure}

    \begin{subfigure}[b]{\textwidth}
        \centering
        \includegraphics[width=0.8\textwidth]{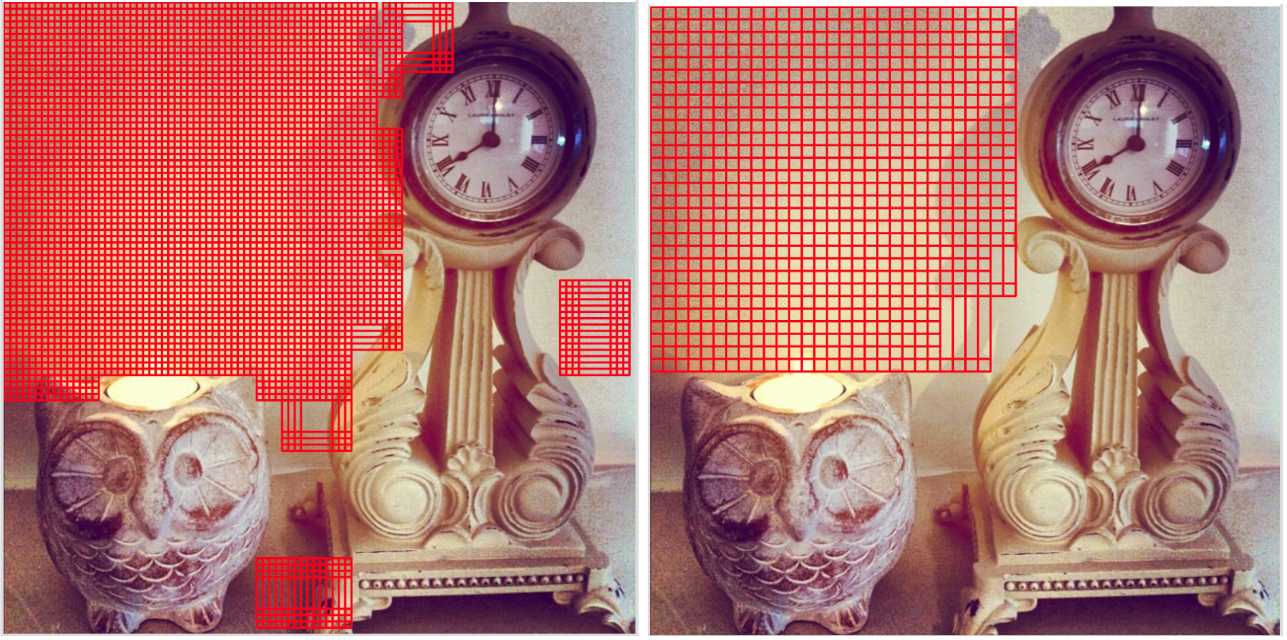}
        \label{fig:fig3}
    \end{subfigure}

    \begin{subfigure}[b]{\textwidth}
        \centering
        \includegraphics[width=0.8\textwidth]{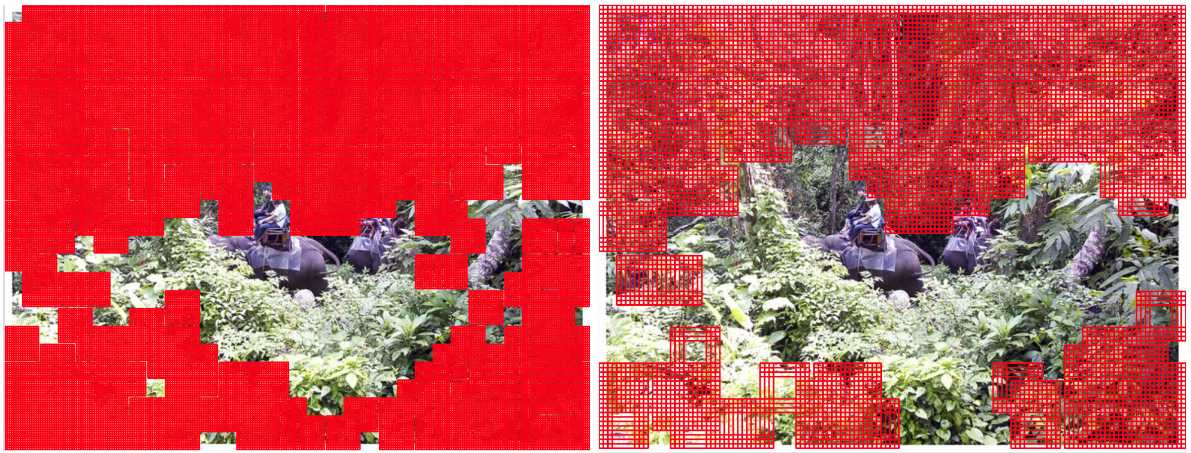}
        \label{fig:fig4}
    \end{subfigure}
    
     \caption{Negative anchorboxes on MS-COCO training set. Here we show negative anchors generated at two scales. For model training 5 scales are used, a common design choice followed in Faster-RCNN \cite{ren2015faster} type architectures.}
    \label{fig:neg_anchor_1}
\end{figure*}

We show examples of negative anchor boxes selected using non-object exemplar set on MS-COCO training images in Figure \ref{fig:neg_anchor_1}. These negative anchors are directly used in training \ours. For a given training image, the negative anchorboxes are selected for all scales used in general Faster-RCNN architecture \cite{ren2015faster, kim2021oln}. In Figure \ref{fig:neg_anchor_1} we show negative anchorboxes computed at two scales for brevity. Selecting negative anchors in multi-scale fashion allows us to accurately localize larger spatial region as non-objects, providing richer supervision during model training. We can observe from the figure that, while the negative anchorboxes donot cover all the non-object regions, it is highly precise in categorizing a region as non-object.

\end{document}